%% file: sample-sigconf.tex
\documentclass[sigconf]{acmart}

\usepackage{multicol}
\usepackage{multirow}
\usepackage{hyperref}
\hypersetup{
    colorlinks=true,
    linkcolor=blue,
    filecolor=magenta,      
    urlcolor=cyan,
    pdftitle={Overleaf Example},
    pdfpagemode=FullScreen,
    }
\AtBeginDocument{%
  \providecommand\BibTeX{{%
    \normalfont B\kern-0.5em{\scshape i\kern-0.25em b}\kern-0.8em\TeX}}}

\copyrightyear{2021} 
\acmYear{2021} 
\setcopyright{acmlicensed}\acmConference[CIKM '21]{Proceedings of the 30th ACM International Conference on Information and Knowledge Management}{November 1--5, 2021}{Virtual Event, QLD, Australia}
\acmBooktitle{Proceedings of the 30th ACM International Conference on Information and Knowledge Management (CIKM '21), November 1--5, 2021, Virtual Event, QLD, Australia}
\acmPrice{15.00}
\acmDOI{10.1145/3459637.3482018}
\acmISBN{978-1-4503-8446-9/21/11}

\def\ourdatasetfull{Trope Understanding in Movies and Animations}
\def\ourdatasetabbr{TrUMAn}
\def\ourmodulefull{Conceptual Storyteller}
\def\ourmodelfull{Trope Understanding and Storytelling}
\def\ourmodelabbr{TrUSt}
\begin{document}

\title{\ourdatasetabbr{}: \ourdatasetfull{}}

\author{
Hung-Ting Su$^{1*}$, Po-Wei Shen$^{1*}$, Bing-Chen Tsai$^{1}$, Wen-Feng Cheng$^{1,2}$, Ke-Jyun Wang$^{1}$, \mbox{and Winston H. Hsu$^{1,3}$}
} \thanks{$^*$ Equal contribution.}
\affiliation{
\institution{$^{1}$National Taiwan University, $^{2}$Microsoft, $^{3}$Mobile Drive Technology}
}

\renewcommand{\shortauthors}{Su et al.}

\begin{abstract}
Understanding and comprehending video content is crucial for many real-world applications such as search and recommendation systems. 
While recent progress of deep learning has boosted performance on various tasks using visual cues, deep cognition to reason intentions, motivation, or causality remains challenging. 
Existing datasets that aim to examine video reasoning capability focus on visual signals such as actions, objects, relations, or could be answered utilizing text bias. Observing this, we propose a novel task, along with a new dataset: \ourdatasetfull{} (\ourdatasetabbr{}) with 2423 videos associated with 132 tropes, intending to evaluate and develop learning systems beyond visual signals. Tropes are frequently used storytelling devices for creative works. By coping with the trope understanding task and enabling the deep cognition skills of machines, data mining applications and algorithms could be taken to the next level.
To tackle the challenging \ourdatasetabbr{} dataset, we present a \ourmodelfull{} (\ourmodelabbr{}) with a new \ourmodulefull{} module, which guides the video encoder by performing video storytelling on a latent space. Experimental results demonstrate that state-of-the-art learning systems on existing tasks reach only 12.01\% of accuracy with raw input signals. Also, even in the oracle case with human-annotated descriptions, BERT contextual embedding achieves at most 28\% of accuracy. Our proposed \ourmodelabbr{} boosts the model performance and reaches 13.94\% accuracy. 
We also provide detailed analysis to pave the way for future research. \ourdatasetabbr{} is publicly available at: \href{https://www.cmlab.csie.ntu.edu.tw/project/trope}{https://www.cmlab.csie.ntu.edu.tw/project/trope}
\end{abstract}





\maketitle
\section{Introduction}
Understanding and comprehending rich information in a video clip is crucial for various applications, including but not limited to information retrieval, recommendation systems, or question answering systems. 
Recent deep learning progress has boosted the performance on many large-scale benchmarks with \textit{shallow visual semantics} such as action recognition, video search, or video question answering (Video QA). 
However, \textit{deep cognition skills} to perform causal and motivational comprehension remains challenging for modern learning models, as mentioned by recent research \cite{bengio2019system,timos}. 

Real-world applications and users might interest in a higher level of concepts beyond shallow visual semantics. For example, audiences of a video clip might be interested in another video with a similar story, plot, or sentiments, such as \textit{bittersweet ending}. Nevertheless, such concepts are rarely demonstrated in visual manners like object or action occurrences. Therefore, a recommendation system needs to understand the causality and the motivation behind \textit{bittersweet ending} where the achievement of the protagonist together with a heavy price paid, instead of tracking shallow semantics, such as a trophy or a crown.
\input{Figures/Figure1}

Many efforts have been devoted to building datasets to evaluate and develop learning systems. Video QA \cite{vdqa,msvdmsrvttqa,anetqa} and multiple-choice movie question answering (MC-MQA) \cite{movieqa,pororoqa,tvqa,violin}, in particular, aim to examine the machine capability of reasoning. However, Video QA datasets focus on visual cues such as actions, objects, or relations, which are too shallow to represent deep cognition skills. MC-MQA datasets, while emphasizing concepts beyond visual cues, such as ``why'' questions, could easily overfit the corresponding language query, as confirmed by recent research \cite{tvqabias,movieqaright,answeraway}. 

We look for another approach: \textbf{Tropes}, which are storytelling devices for creative works such as movies, animations, or literature. Beyond object and action co-occurrences, they are the tools that art creators use to deliver abstract and complex ideas to the audience without spelling out all the details. For example, \textit{Bad Boss} means a boss callously mistreats their employees. This trope could be portrayed with a scene where a boss is punching or even killing their subordinates. While this concept is trivial for an educated human, it requires a learning system to comprehend the motivation of the action and the cause of the torture scene. Hence, different from conventional Video QA tasks, tropes involve consciousness, systematic generalization, causality, and motivational inference \cite{timos}. A recent work \cite{timos} utilizes movie synopses and tropes from the TVTropes database to evaluate the reasoning capability of learning systems. However, movie synopses are written by humans and consequently rarely available in practice. Furthermore, movie synopses usually contain implicit human interpretation instead of raw visual signals. Therefore, it would be easier for a machine to capture the tropes in films using human-written synopses.

We are optimistic that trope understanding capability could bring significant leaps forward in both data mining applications and algorithms, from developing a recommendation system to studying motivational behaviors beyond visual semantic. 
Therefore, we propose a novel task, along with a new dataset, \ourdatasetfull{} (\ourdatasetabbr{}), including 2423 videos associating with 132 tropes. \ourdatasetabbr{} inputs video and audio signals, reflecting the real world where people interact with each other and the environment instead of human-written synopses. Unlike traditional datasets and approaches, which focus on visual signals, our \ourdatasetabbr{} requires deep cognition skills of learning systems. For instance, as shown in Figure \ref{fig:test}, similar visual clues (watching a screen) are portrayed in the first and the second columns but delivering different stories. In contrast, the video clip in the third column has the same trope as the second column (Bad Boss), conveying similar abstract concepts, but the visual contents are totally different.

To tackle the novel yet challenge \ourdatasetabbr{}, we propose a new \ourmodelfull{} (\ourmodelabbr{}) model, which jointly understand trope and perform storytelling on a latent space in a multi-task manner. Specifically, the \ourmodulefull{} generates a story embedding vector, which represents the video description. Next, the generated vector is optimized by a human-written description embedded by a pre-trained text encoder. Finally, the generated story embedding vector is fed to a trope understanding model to determine the output trope. By utilizing the generated story embedding, human-written descriptions are not needed during inference.

Experimental results demonstrate that modern learning systems still struggle to solve the trope understanding task, reaching at most 14\% accuracy. State-of-the-art models, including graph-based L-GCN Video QA model \cite{L-GCN2020AAAI}, and cross-modal pre-training-based XDC action recognition model \cite{xdc}, while utilizing visual semantics to perform well on existing tasks, could not solve the trope understanding task. With the aid of human-written description, the accuracy could be boosted to 28\%, indicating that trope understanding dataset using movie synopses \cite{timos} might over-estimate machine deep cognition capability. Moreover, we provide a comprehensive analysis to pave a new path for future research. Consequently, we are optimistic that our proposed task and dataset could bring learning systems to the next level.


\section{Impact and Potential Extensions}
This section discusses the impact and several potential new tasks or applications on the basis of our \ourdatasetabbr{} dataset. Apart from solving our challenging task and dataset, future work might be interested in extending our work to build a new dataset or an application.

\paragraph{Search and Recommendation} Retrieving web content based on queries or watching history is essential for various web applications. A user might seek content that he/she is interested in based on cues beyond shallow textual or visual semantics. For example, looking for a \textit{bittersweet} movie. Our \ourdatasetabbr{} provides a test-bed to examine learning systems' capability of search and recommendation. For example, future research could utilize our trope annotations and categories to formulate a trope-based video recommendation task, i.e., recommending a video based on another video with the same or a similar trope. 

\paragraph{Trope-based Video Description Generation} Summarizing a video or a document with natural language sentences is a crucial task that has been studied for years.
Conventional benchmarks such as MSVD \cite{msvd} or MSRVTT \cite{msrvtt} mostly focused on captioning a video based on action signals. As our \ourdatasetabbr{} presents both videos and corresponding descriptions, future work could leverage these video-description pairs to learn to generate descriptions based on video clips.

\paragraph{Disentangling Motivation behind Actions} Some significantly dissimilar tropes are portrayed in similar actions or events. For example, \textit{asshole victim} and \textit{heoric sacrifice} are sharply different but could both displayed by ``someone's death" in a video clip or a novel. By using these tropes and associated videos in \ourdatasetabbr{}, future study might want to explore disentangling deeper cognition such as motivation from video representation and develop downstream applications.

\section{Related Work}
\paragraph{Video QA Datasets}Video QA datasets were widely used to evaluate machine capability of understanding a video. Early Video QA works such as \citet{vdqa} and \citet{msvdmsrvttqa} leveraged existing video datasets with captions and annotated question-answer pairs using a text question generation tool \cite{HnS}. The assumption behind these datasets is that a machine needs to understand the video content in order to answer a question. In these datasets, a set of answers (usually around 1,000) was pre-defined and classified into several categories. Most answers in Video QA datasets are an entity (e.g., dog) or an action (e.g., dancing). Therefore, Video QA datasets could be narrowed down to action and object recognition tasks according to a text query to some extent. 
\citet{anetqa} proposed a human-annotated Video QA dataset, Activitynet-QA (Anet-QA), and extended the question types to include color, location, and spatial and temporal relations. However, Anet-QA did not incorporate causal and motivational queries and therefore could not examine the machine capability of deep cognition skills. 
\input{Figures/categorycloud}

\paragraph{Movie Understanding (MU) Datasets}
MU datasets, while shared some properties with Video QA datasets, focused more on deeper reasoning capability (e.g. ``why'' questions). Most MU datasets were formed as multiple-choice movie question answering (MC-MQA) and properly designed distractor options to examine the machine reasoning capability. MovieQA \cite{movieqa} labeled 15,000 multi-choice questions associated with 400 movies. While opening the research of movie understanding, the main drawback of MovieQA is that questions are labeled using plot synopses instead of movies themselves. TVQA \cite{tvqa} collected 6 TV series and annotated 100,000 multiple-choice questions according to the videos. TVQA dataset mainly focused on temporal relations (i.e. All questions consisted ``before'', ``after'', or ``when''). VIOLIN \cite{violin} proposed a Video-and-Language Inference task where positive-negative statement pairs were provided, and the model was asked to determine which one was correct. While MU datasets provided deeper questions to evaluate the machine reasoning capability, recent research \cite{movieqaright,tvqabias,answeraway} suggested that models tended to overfit language queries (questions or language inference). Therefore, a learning model might reach a high score by utilizing bias instead of understanding movie contents. Our dataset, In contrast, requires the model to process raw signals to perform the trope understanding task.

\paragraph{Trope Understanding}
Tropes were introduced to the multimedia community by \citet{HarnessingAI}. Recently, \citet{timos} proposed a Trope in Movie Synopses (TiMoS) dataset with about 6,000 movie synopses from MPST dataset \cite{mpst} and 95 associated tropes. Different from Video QA datasets, the trope dataset aimed to examine the machine capability of deep cognition, including but not limited to consciousness, systematic generalization, causality, and motivation. As trope detection tasks do not require additional queries, machines cannot capture bias in language queries. However, the inputs of TiMoS are movie synopses instead of movies themselves. On the other hand, our \ourdatasetabbr{} directly inputs movie contents, including video and audio, which fits real-world scenarios such as search or recommendation.

\section{\ourdatasetabbr{} Dataset}
\subsection{Overview} \label{section:dataset:overview}

We present a novel dataset \ourdatasetabbr{} (\ourdatasetfull{}) which includes 2423 videos with audio and 132 tropes. We also include human-annotated video descriptions in our dataset to compare the domain gap between raw visual and audio signals and human-written text. We classify tropes into 8 categories by their properties. As categories are not orthogonal, some tropes belong to 2 or more categories. Figure \ref{fig:cate} shows the trope clouds for each category.

\textbf{Character Trait} tropes focus on a specific role and their characteristic. The trait would be portrayed with their behavior instead of a direct description, such as  \textit{Big Bad}, which shows someone in the video with evil plans and causes all the bad things to happen.

\textbf{Role Interaction} tropes describe the actions, conversations, or encounters of roles in the video. \textit{Bad Boss} is when a boss is being mean to their employee. However, a bad boss might be a good father or even a hero.

\textbf{Scene Identification} tropes focus on specific views or objects in a scene. This category could not be solved with only object co-occurrences. For example, \textit{playing with fire} where a character is able to control fire for their utility, could not be simplified as ``someone occurs with fire''. 

\textbf{Situation understanding} tropes depict a short-term scenario where there are some events happening. These events could be composed of certain entities, objects, actions, or conversations and convey some information or concepts to the audience. An example is \textit{Berserk Button}, which represents a character flies into a rage by minor or generally insignificant thing. Detecting this trope requires a model to understand the motivation that triggers someone's anger.

\textbf{Story Understanding} tropes describe a long-term scenario of the video, usually combined by multiple situations. These tropes need to fully understand what is happening in the video and realize the conversation's meaning. e.g. \textit{``The Reason You Suck'' Speech} is a character delivers a speech to another character about why he sucks.

\textbf{Sentiment} tropes deliver some emotion by elements in a video, includes but not limited to the scene, conversations, speech, or music. These tropes need to realize the emotions that videos convey to the audience, e.g. \textit{Downer Ending} is a movie or TV series that ends things in a sad or tragic way, the scene of the videos usually becomes gloomy and the music is often melancholy.

\textbf{Audio} tropes focus on audio, the music tune in the video, the speech or conversation content, or the tone of the speakers. These tropes are hard to classify with only video appearance for humans. Such as \textit{Villain Song}, knowing what is the character singing in the video and the tune of the music make us identify the trope easier.

\textbf{Manipulation} tropes are tropes where the director uses different photography skills (\textit{Running Gag}) or script (\textit{Shout Out}) to interact with the audience. This type of tropes strongly needs deep cognition skills, having the knowledge or related concepts may help to recognize them.

\subsection{Data Collection}
\paragraph{Trope and Video collection}
We collected tropes and videos from a Wikipedia-style database, TVTropes\footnote{https://tvtropes.org/}. Each trope is along with the definition, several example videos, and related video descriptions, where these data are annotated by web users. Specifically, we query TVTropes for the example videos from each trope and also the description of the videos.

\paragraph{Trope selection}
After video collection, we get more than 10k videos and about 4k different tropes. Since most of the tropes have only a few video examples, we select the most frequent tropes from the data and get 132 tropes and 2423 video examples at last, where each trope has more than 10 examples. Finally, we split the dataset with 2423 videos into 5 splits (495/487/478/483/480, 20.43\%/20.10\%/19.73\%/19.93\%/19.81\%). In order to composed a 5-fold cross-validation with validation and test set, we further split 12.5\% of data from training set (10\% of whole data) as validation set. Thus, each training process has training, validation, and test set with a size ratio 7:1:2.
\subsection{Data Analysis} \label{section:dataset:analysis}
\input{Figures/statisticcate2}
\input{Figures/occurence}
Table \ref{tab:statistic2} summarizes the video tropes statistic in each category. For the whole dataset, the average length of videos is about 1 minute, but the standard deviation of the dataset is quite large, showing that videos in \ourdatasetabbr{} dataset are very diverse. The last column of Table \ref{tab:statistic2} shows the number of tropes. 34\% of tropes occur in multiple categories, which shows that tropes can be understood from different aspects. In Table \ref{tab:occurrence}, the distribution of different lengths of tropes' videos shows that about 14\% of videos are long (over 2 minutes), and about 17\% of videos are short (less than 20 seconds). The variety length of videos makes trope detection harder.

\input{Figures/HumanRes}
\subsection{Human Evaluation on \ourdatasetabbr{}}\label{sec:humaneval}
To better understand the collected dataset and provide directions for future research, we conduct a human evaluation on \ourdatasetabbr{}. We sample 100 video examples for human evaluation where each human tester was asked to select a trope in 5 trope options. 4 distractor candidates are randomly selected from the rest of the tropes, including 2 in the same category as the answer to make the evaluation more challenging. Note that human annotators in the evaluation are not experts, so human evaluation errors do not indicate an unanswerable example.

\paragraph{Overall comparison} Table \ref{tab:humanres} shows non-expert human evaluation results. Overall, visual signals play a relatively more important role comparing to audio signals. Without watching the movie (Audio Only), humans could achieve 54.0\% accuracy, while the Video Only could lead to 69.0\% accuracy. Combining both modalities (Visual+Audio) enhances the human performance to 77.0\%, indicating that fusing information from both visual and audio is crucial for understanding a trope.

\paragraph{Trope modality}
There are some examples that require a specific modality. For example, many audio category tropes centering on music or dialogues could hardly be recognized with video only. On the other hand, several manipulation category tropes could rarely be detected without videos, such as \textit{animation bump} or \textit{overly long gag}. Therefore, using merely a single modality cannot understand all tropes. 

\paragraph{Conceivable Tropes}
Some tropes seem require a specific modality. However, human could conceive the trope with another modality. For example, a video with \textit{villain song} could be conceived by watching a villain-like character singing. Additionally, some visual tropes such as \textit{groin attack} could be conceived with the sound of punches and screaming.

\paragraph{Complement Modalities}
Several examples require both audio and video to understand, such as a \textit{Screw this, I'm outta here!}. With only video available, the human tester misunderstand the clip as \textit{Big NO!} as the video shows a shot of a person seems screaming. On contrary, with only audio, another human test labeled it as \textit{Feud Episode} because it sounds like two people are feuding with each other to break up. With both video and audio, a human annotator could correctly get the trope. This suggests that fusing audio and visual features might be a way to tackle with trope understanding task.

\paragraph{External Knowledge}
Certain tropes such as \textit{getting crap past the radar} are represented in more obscure ways because they could violate some censorship standards. As the trope is portrayed with some allusions in literature, history, or memes, readers need to externally understand the allusions in order to comprehend the trope. We observe that this kind of tropes is where human annotators failed with accessing both video and audio. Intuitively, it would also be challenging for machines because it requires machines to learn from very specific knowledge sources.

\subsection{Data Availability}
The \ourdatasetabbr{} homepage\footnote{https://www.cmlab.csie.ntu.edu.tw/project/trope} provides a brief introduction of \textit{Trope} and the features and the usage of our dataset. We also display some samples on the page for new researchers to acquaint them with this novel and intriguing task. The data we provide includes:
\begin{itemize}
    \item \ourdatasetabbr{}:  \ourdatasetfull{} dataset has five-fold data split files, each split file includes train, validation, and test data. Each example is associated with its video ID, trope name, human-annotated description, and the detected ASR results.
    \item Visual features: We provide ResNet-101 \cite{resnet} and  S3D \cite{s3d} features for future researchers, the usage is introduced in our page.
    \item Audio features: We also provide SoundNet \cite{soundnet} features for multi-modal reasoning.
\end{itemize}

\input{Figures/cikm_model}

\section{\ourmodelfull{} (\ourmodelabbr{}) Model}\label{sec:model}
To take the first step to tackle the challenging \ourdatasetabbr{}, we propose a new \ourmodelfull{} (\ourmodelabbr{}) network with three modules: (1) Video Encoder (Section \ref{sec:model:encoder}), which encodes a multi-modal video clip into a video embedding vector. (2) \ourmodulefull{} (Section \ref{sec:model:storytelling}), which generates a story embedding vector according to the video embedding. The story embedding vector is optimized by minimizing the distance to video description vector encoding by a pre-trained contextual embedding model (e.g., BERT). (3) Trope Understanding (Section \ref{sec:model:trope}), which performs trope classification based on video embedding and story embedding vectors.

\subsection{Video Encoder}\label{sec:model:encoder}
The video encoder module takes N-stream inputs $\{F^{1}, F^{2} \cdots F^{N}\}$ where each stream represents a modality such as visual or audio features, and outputs a video embedding $V$. 
\begin{equation}
    V = VideoEncoder(\{F^{1}, F^{2} \cdots F^{N}\}, \theta^{E})
\end{equation}
, where $\theta^{E}$ are trainable parameters.

 Note that the module is flexible with additional features and different encoder architectures. In this work, we use visual, audio, ASR, and object features (See \ref{sec:exp:modality} for details). For the video encoder, we leverage and slightly modify previous work \cite{violin,L-GCN2020AAAI} for video-and-language inference and video question answering. First, in each stream, the input feature $F^{i}$ is encoded with a per-modality (e.g., audio) encoder:
 \begin{equation}
    X^{i} = Encoder^{i}(F^{i}, \theta^{{E}_i})
 \end{equation},
 where $X^{i}$ is encoded $F^{i}$ and $\theta^{{E}_i}$ stands for trainable parameters. 
 
 
 Afterward, we concatenate all encoded features:
 \begin{equation}
     V = [X^{1};X^{2} \cdots X^{N}]
 \end{equation}
 The encoded video embedding $V$ is then utilized to represent the input video.

\subsection{\ourmodulefull{}}\label{sec:model:storytelling}
We design a novel conceptual storyteller to guide the model by leveraging video descriptions. The intuition behind the module design is two-fold. First, by learning to tell a story, the video encoder could receive further signals from the video description. Second, the model-generated story is then fed into a trope understanding module to augment the video embedding without the need for human-written descriptions during inference.


Given a video embedding vector $V$, the module generates a \textit{story embedding} vector:
\begin{equation}
  S = ConceptualStoryteller(V, \theta^{S})  
\end{equation},
where $\theta^{S}$ refers to trainable parameters. Instead of actually generating the story by tokens (i.e. video captioning), we generate a story embedding vector on a latent space for two reasons. (1) A vanilla encoder-decoder model requires an argmax operation to generate a token and the gradient could not be back-propagated. (2) Comparing to conventional video-to-text tasks and datasets, the word distribution of video descriptions is much sparser. Furthermore, we have much fewer examples available. 

To optimize the generated story embedding vector, we leverage human-written descriptions and a freezed,  pre-trained contextual encoder to obtain contextual embedding:
\begin{equation}
    C = ContextualEncoder(D)
\end{equation}
, where we utilize BERT-base \cite{bert} in this work. 

Then, we minimize the story loss, the distance between generated story embedding $S$ and embed descriptions $C$:
\begin{equation}
    L^{S} = 1 - Cosine(S, C)
\end{equation}
, where we use cosine similarity as our distance function.

\subsection{Trope Understanding}\label{sec:model:trope}
The trope understanding module utilizes video embedding and our generated story embedding to predict a trope. First, we generate a trope distribution $T^{pred}$:
\begin{equation}
    T^{pred} = TropeUnderstanding(V, S, \theta^{U})
\end{equation}
, where $\theta^{U}$ are trainable parameters.

Then, we apply a cross entropy for trope loss on $T^{pred}$ and ground truth $T^{gt}$:
\begin{equation}
    L^{T} = CrossEntropy(T^{gt}, T^{pred})
\end{equation}

Finally, we weighted sum the losses together, where $\alpha$ and $\beta$ is pre-defined hyper-parameter:
\begin{equation}
    L = \alpha * L^{S} + \beta * L^{T}
\end{equation}

\input{Figures/Results_new}
\section{Experiments}
\subsection{Modality}\label{sec:exp:modality}
\paragraph{Visual}
For visual signals in videos, we extract static appearance and dynamic motion features in a video to evaluate how the static and dynamic information affect trope understanding.
Specifically, we apply ResNet pre-trained on ImageNet image classification to extract appearance features, and use S3D pre-trained on Kinetics action recognition to extract motion features. 
We denote appearance features as $F^{a} = [f^{a}_{1}, f^{a}_{2}, f^{a}_{3},\cdots,f^{a}_{N}]$, and motion features as $F^{m} = [f^{m}_{1}, f^{m}_{2}, f^{m}_{3},\cdots,f^{m}_{N}]$, in which N is the number of frames. In our experiments, we set N to 100 and truncate the frames longer than it.\footnote{Only 0.1\% of videos in \ourdatasetabbr{} (2 in 2423) are truncated.} Both ResNet and S3D features are extracted with 0.5 fps, the dimensions of the features are 2048 and 1024. 
\paragraph{Audio}
For audio features, we use (1) SoundNet to encode all the sound that appears in the video, includes music, speech, natural sound, etc. The feature denotes as $F^{aud} = [f^{aud}_{1}, f^{aud}_{2}, f^{aud}_{3},\cdots$, $f^{aud}_{K}]$, K is the length of an audio signal, and the feature dimension is 1024. (2) ASR model from Google Cloud Speech to Text API to extract speech transcript in the videos, and use BERT \cite{bert} to encode the transcript. We denote the feature as $F^{asr} = [f^{asr}_{1}, f^{asr}_{2}, f^{asr}_{3},\cdots$, $f^{asr}_{L}]$, L is the length of the speech, and the feature dimension is 768.
\paragraph{Object}
L-GCN \cite{L-GCN2020AAAI} requires local object features to input the graph-based model. We use Faster-RCNN \cite{frcnn} to extract object features, where Faster-RCNN was pre-trained on Visual-Genome. The object set $R = \{o_{n, k}, l_{n, k}\}^{n=N,k=K}_{n=1,k=1}$, o is the \textit{k} th object detected at \textit{n}-th frame with dimension 2048, and l is the spatial location of each object. Each frame has K detected object and the total length of the frame is N.

\subsection{Compared Methods} \label{section:experiment:methods}
We examine trope understanding capability of modern learning systems on our \ourdatasetabbr{} dataset, including a modified VIOLIN \cite{violin} model as our baseline, a modern GCN based Video QA model, L-GCN \cite{L-GCN2020AAAI}, and a state-of-the-art self-supervised cross-modal pre-training method, XDC \cite{xdc}. To reveal gaps from machine to human, we also evaluate (1) Oracle model using human-written descriptions and (2) Human performance with sampled examples.

\paragraph{Baseline}
Follow previous work \cite{violin, tvqa}, we use LSTM network to encode visual and audio features. Since trope detection is different from Video QA and video inference problems, we remove the question (query)-video attention block in \cite{violin} as our baseline.

\paragraph{L-GCN}
L-GCN \cite{L-GCN2020AAAI} is a location-aware graph-based model, where it models the relation between objects in all the video frames, and shows the power on video QA tasks. We remove the QA block in the model and adjust it for our task.

\paragraph{XDC}
XDC \cite{xdc} is the state-of-the-art self-supervised method that leverages video and audio signals in the video for action recognition and audio classification tasks. We use the released visual model pre-trained with IG-Kinetics dataset.

\paragraph{Oracle}
Different from those raw signals we extracted from the videos, we have collected the human-annotated video descriptions. These descriptions directly point out the most important part of the videos and act as a guide to realize the tropes instead of through the videos. For these descriptions, we use BERT as a feature extractor to extract the feature from these descriptions as how we preprocess the ASR video transcript. These features are pass to our baseline model that we can easily compare the capability of raw signals and oracle annotations.

\subsection{Results and Discussion} \label{section:experiment:results}
\paragraph{Modality} The first to fourth rows of the first block of Table \ref{tab:exp} shows the single-modal performance of the baseline model. All variants reach at best 11.60\% accuracy, indicating that trope understanding is a challenging task. Also, the visual model ($9\sim11\%$ accuracy) has a better performance compared to the audio model ($4\sim5\%$), even in the Audio category. This echos the human evaluation in Section \ref{sec:humaneval} where visual signals could play a slightly more crucial role in the task. Additionally, audio signals could either be too sparse (sound) or suffer information loss (ASR). 
The ASR model, while ignores music, performs better than the sound model, especially in the scene identification (third) category where scene-related concepts might be mentioned in the dialogues. The fifth and sixth rows of the first block demonstrate multi-modal baseline model performance. Fusing visual and ASR features generally improves the accuracy to 12.01\%, revealing that modalities are complementary. 

\paragraph{Existing state-of-the-art method capability}
The second block of Table \ref{tab:exp} shows the performance of L-GCN \cite{L-GCN2020AAAI}, a state-of-the-art Video QA model using detected object features. The model has the accuracy between audio models and visual models, revealing that the composition of detected objects does not represent tropes well. 
The third block shows XDC \cite{xdc}, a self-supervised audio-video clustering approach that reaches state-of-the-art on action recognition tasks. The model performs significantly better on the audio category despite only inputting visual signals, which shows that cross-modal audio-visual pre-training could help the model to capture audio-related deep semantics in the video, and also echos the conceivable tropes we mentioned in human evaluation. However, it does not perform well overall (7.35\% accuracy), revealing that generally transferring visual semantics to trope understanding remains challenging.

\paragraph{TrUSt}
 The fourth block of Table \ref{tab:exp} shows the results of our \ourmodelfull{} model: \ourmodelabbr{}. As shown in the first row, our \ourmodulefull{} component boosts the baseline model to 12.88\% accuracy. The performance of most categories rises, especially the character trait, role interaction, and audio. 
 It is reasonable as the character trait and the role interaction require understanding role characteristics or intentions behind their actions. Therefore, storytelling helps the model to apprehend them. 
 Also, audio performance is improved because the story embedding complements raw signals. Specifically, without \ourmodulefull{}, the model might get that the video is audio-related but does not comprehend the exact one. i.e., misunderstanding a ``Title Theme Tune" as a ``Villain Song". The second row demonstrates the performance of \ourmodelabbr{} with an additional frcnn stream. We further boost the performance to 13.94\% accuracy, which is the best among all non-oracle models. \ourmodelabbr{} with only S3D+ASR drops the performance in scene identification, which might stem from the story embedding with sparse raw signals instead of specific objects. With the aid of frcnn stream, the story embedding could be generated with both sparse visual (the whole frame) and dense object cues and enhances the scene identification performance to 15.93\%, which is also better than multi-modal baseline or L-GCN model alone, and indicates the effectiveness of the proposed \ourmodulefull{} module.

\paragraph{Oracle model w/ human-written description}
It worth noting that the oracle model (sixth block) which accesses the human-written description gets 27.84\% accuracy. While it is still far from human performance, the score is more than doubled of the best baseline model. Therefore, we argue that trope understanding in movies and animations is much more difficult than previous proposed synopses trope understanding \cite{timos}, and could examine the machine capability of processing raw signals, which is crucial for real-world applications such as movie recommendation systems. 

\input{Figures/ablation}
\paragraph{Ablation study on \ourmodelabbr{}}
Table \ref{tab:ablation} demonstrates the ablation study on \ourmodelabbr{} to indicate the effectiveness of each proposed module. We compare (1) \ourmodelabbr{}, (2) \ourmodelabbr{} without feeding story embedding back to trope understanding module (multitask structure only), and (3) the baseline methods. The first row of Table \ref{tab:ablation} displays the result with visual and audio signals, and the second row shows the result with visual, audio, and object signals. The second column shows a comparable result to the baseline model without (first row) and 1.0 performance gain with object signals (second row). It demonstrates that multitasking alone guides the video encoder with adequate input signals. Also, as shown in the third column, the full \ourmodelabbr{} further boosts $0.9\sim1.5$ accuracy by leveraging the story embedding.

\input{Figures/Cases}
\paragraph{Qualitative Analysis} \label{para:analysis}
Figure \ref{fig:cases} demonstrates some cases predicted by the oracle model, \ourmodelabbr{}, the baseline model, LGCN, and XDC. The first and second cases show the correct answer predicted by \ourmodelabbr{} and the oracle model; the third case is the failure case of \ourmodelabbr{}. 
In the first case, the oracle model understands the trope based on the human-written description, indicating that descriptions could hint at the model and simplify trope understanding. Our \ourmodelabbr{} also comprehends the video with the aid of \ourmodulefull{} module.
Simultaneously, the baseline and XDC somehow understand the sad mood portrayed in the film. However, they misunderstand that Emma reminisces about the old days as someone sacrificed heroically. 
The second case is an expository theme of the video. The oracle model and \ourmodelabbr{} perform well in this case. These two cases demonstrate the effectiveness of our proposed \ourmodelfull{} model, which leverages story embedding without needing human-written descriptions during inference. 
Concurrently, the baseline model, LGCN, and XDC, despite capturing the musicality of the video, fail to apprehend the trope.
In the last case, we can observe that only the baseline method gets the correct answer. To predict \textit{Shock and Awe}, it is crucial to capture the cues of light or the special effects of the weapon provided by the raw visual and audio signals. The oracle model and \ourmodelabbr{} might over-conceive the descriptions (story embedding) and output a wrong answer. Observing this issue, future work might want to investigate the border of storytelling and over-conceiving to further improve the performance.

\section{Conclusion}
We propose a novel task, along with a new dataset, 
\ourdatasetabbr{}. 
Different from existing datasets, \ourdatasetabbr{} requires deep cognition skills to comprehend causality and motivation beyond visual semantics, and could not be solved with language bias and implications. \mbox{By developing machines with trope understanding capability}, various data mining applications and algorithms could be taken to the next level. 
To tackle this challenging task, we present a new model, 
\ourmodelabbr{}, 
which jointly perform storytelling and trope understanding with a novel module \ourmodulefull{}. Experimental results show that learning systems that perform well on conventional video comprehension benchmarks reach at most 12.01\% accuracy, revealing the room for improvement for modern learning systems. Our proposed \ourmodulefull{} boosts the model performance by 2\% accuracy and reaches the state-of-the-art performance of 14\% accuracy. 
Additionally, the oracle case that using human-written description instead of raw signals could boost machine performance to 28\% of accuracy, indicating that our dataset is more challenging than trope understanding dataset with movie synopses. Therefore, we believe that our proposed task and dataset could lead to a new path for future research and applications. 

\section*{Acknowledgement}
This work was supported in part by the Ministry of Science and Technology, Taiwan, under Grant MOST 110-2634-F-002-026. We benefit from NVIDIA DGX-1 AI Supercomputer and are grateful to the National Center for High-performance Computing.

\clearpage
\bibliographystyle{ACM-Reference-Format}
\bibliography{sample-sigconf}
\end{document}

%% file: Figures/Figure1.tex
\begin{figure*} 
\centering
\includegraphics[width=\textwidth]{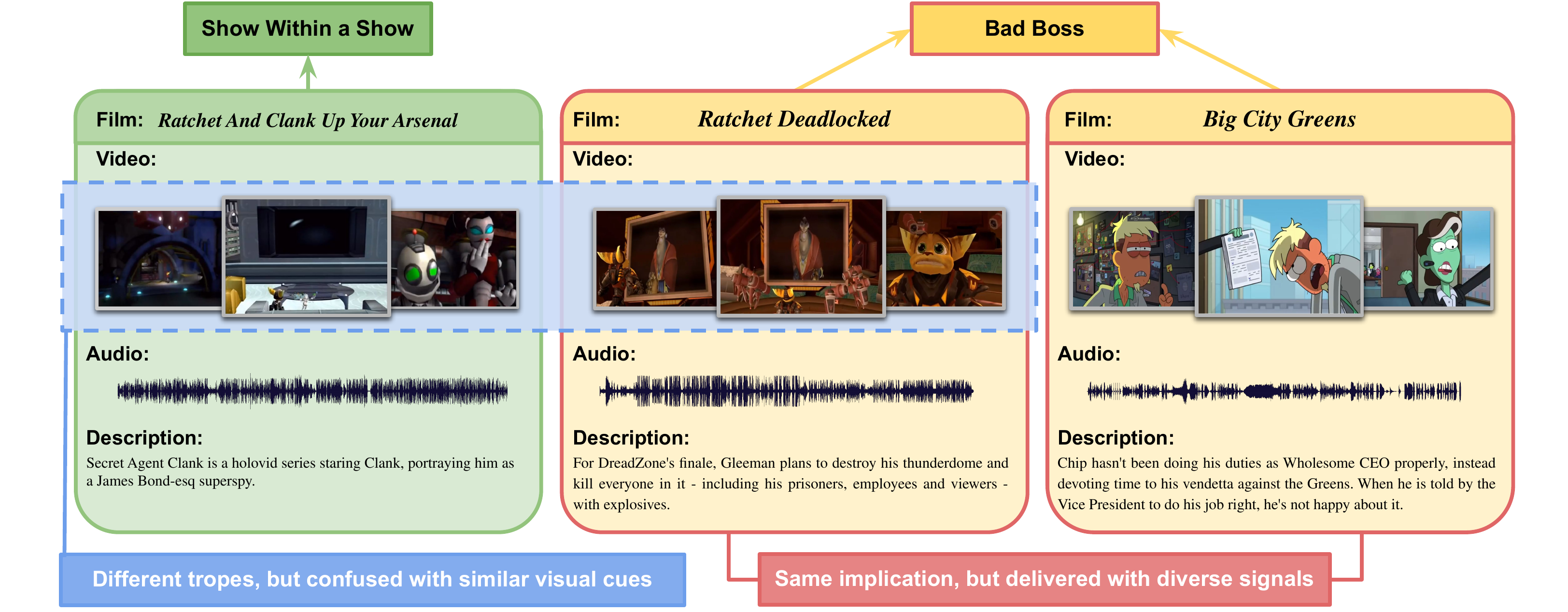}
\caption{
{{\bfseries \ourdatasetfull{} (\ourdatasetabbr{}). Trope understanding requires deep cognition skills to comprehend the causality and the motivation beyond visual semantics. The first column and the second column are represented in similar visual semantics (watching TV) but conveying different stories. On the other hand, the third column is visually different from the second column, delivered with diverse signals, but is more similar in human cognition.
}
}
{ 
}
}
\label{fig:test}
\end{figure*}

%% file: Figures/categorycloud.tex
\begin{figure*} 
\includegraphics[width=\textwidth]{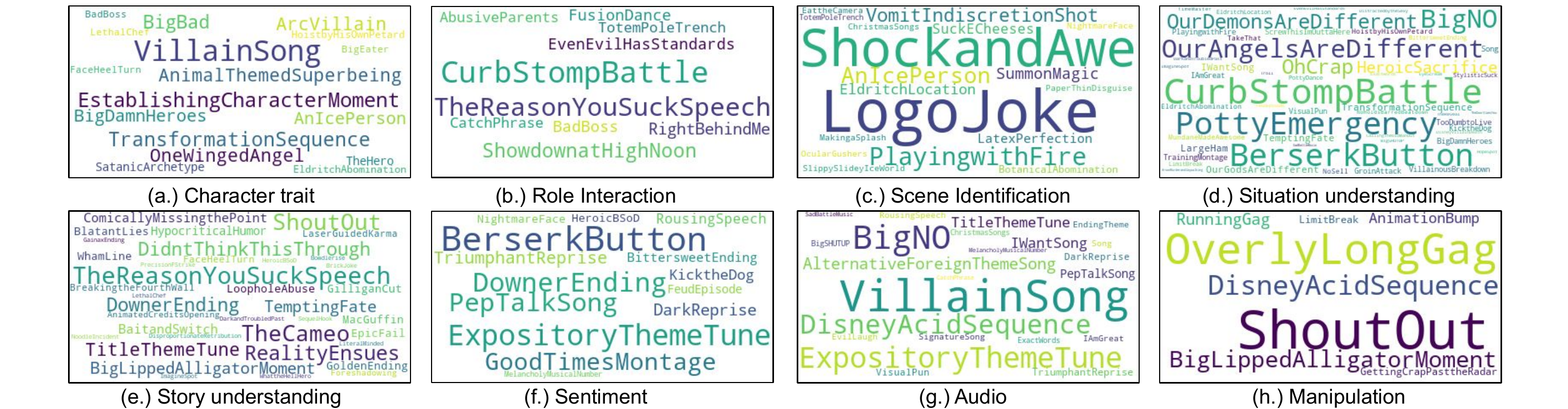}
\caption{
{{\bfseries Word cloud of trope categories.} Size of words are proportional to the frequency of tropes in the new collected dataset \ourdatasetabbr{}. (See Section \ref{section:dataset:overview})
}
}
\label{fig:cate}
\end{figure*}

%% file: Figures/statisticcate2.tex
\begin{table}[ht]
    \centering
    \noindent
        \begin{tabular}{l|ccccc|c}
        \toprule
        Category & Avg. & Median & Min & Max & \(\sigma\) & Number \\
        \midrule
        C. Trait & 77.86 & 71.00 & 4.33 & 237.67 & 43.95 & 17 \\
        R. Inter. & 69.53 & 57.00 & 4.67 & 157.00 & 42.33 & 10 \\
        Scene Id. & 49.60 & 36.67  & 3.67 & 150.00 & 39.24 & 18  \\
        Story. U & 55.51 & 45.7 & 3.67 & 237.67 & 38.95 & 37 \\
        Situ. U & 57.43 & 45.67 & 3.67 & 233.67 & 42.03 & 51  \\
        Sent. & 75.93 & 68.33 & 4.00 & 233.67 & 41.08 & 14 \\
        Audio & 79.39 & 72.00 & 4.33 & 178.00 & 46.10 & 22 \\
        Mani. & 61.34 & 51.00 & 6.33 & 156.67 & 44.09 & 8 \\
        \midrule
        All & 61.90 & 52.00 & 2.33 & 237.67 & 42.71 & 132 \\
        \bottomrule
        \end{tabular}
    \caption{{\bfseries\ourdatasetabbr{} dataset video statistics.} It shows the statistic of video length (in second) and the number of tropes in each category. The average video length is long, and the standard deviation is quite large, shows that videos in \ourdatasetabbr{} dataset are very diverse. 34\% of tropes belong to multiple categories, which shows that tropes can be recognized from different aspects.  (\(\sigma\) denotes standard deviation.) (See Section \ref{section:dataset:analysis})}
    \label{tab:statistic2}
\end{table}

%% file: Figures/occurence.tex
\begin{table}[ht]
    \centering
    \noindent
        \begin{tabular}{c|ccc}
        \toprule
        Time & Short(< 20 sec) & Median & Long(> 2 min) \\
        \midrule
        (\%) & 17.62 & 68.01 & 14.37  \\
        \bottomrule
        \end{tabular}
    \caption{{\bfseries The statistic of video tropes' occurrence in percentage.}) (See Section \ref{section:dataset:analysis})}
    \label{tab:occurrence}
\end{table}

%% file: Figures/HumanRes.tex
\begin{table}[]
    \centering
    \begin{tabular}{ccc}
    \toprule
         Visual Only & Audio Only & Visual+Audio 
         \\
         69.0 & 54.0 & 77.0 
         \\
    \bottomrule
    \end{tabular}
    \caption{Human and machine evaluation result with sampled subset with 100 examples. (Section \ref{sec:humaneval})}
    \label{tab:humanres}
\end{table}

%% file: Figures/cikm_model.tex
\begin{figure*} 
\centering
\includegraphics[width=\textwidth]{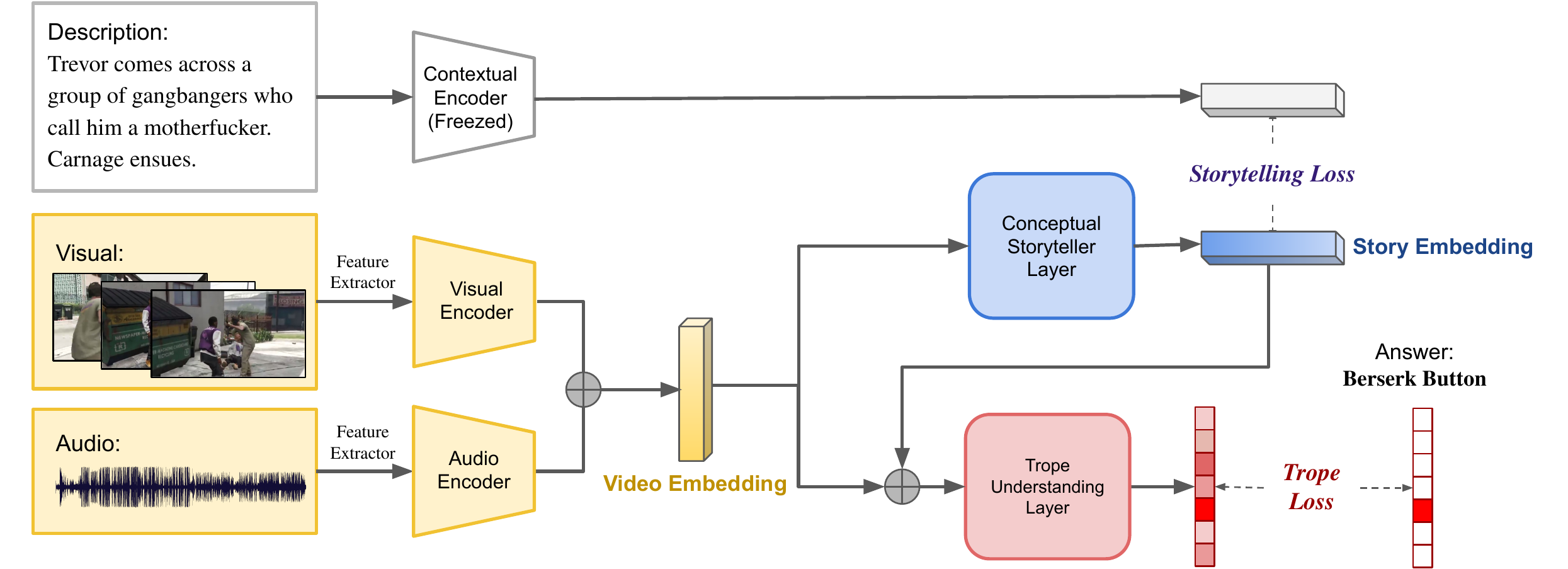}
\caption{
{{\bfseries Proposed \ourmodelabbr{} model. \ourmodelabbr{} model consists of a multi-modal video encoder module (yellow, left, Section \ref{sec:model:encoder}), our proposed novel \ourmodulefull{} module (blue, top right, Section \ref{sec:model:storytelling}), and a trope understanding module (red, bottom right, Section \ref{sec:model:trope}). First, the Video Encoder module takes multi-modal inputs and encodes a video embedding. Next, the \ourmodulefull{} generates a story embedding according to the video embedding to provide further information for trope understanding without the requirement of human-written descriptions during inference. Furthermore, story embedding generation guides the video encoder with additional signals. Finally, the trope understanding module predicts the trope according to input video embedding and generated story embedding.
}
}
{ 
}
}
\label{fig:model}
\end{figure*}

%% file: Figures/Results_new.tex
\begin{table*}
  \centering
  \begin{tabular}{cccc|c|cccccccc}
    \toprule
    \midrule
     Method & \multicolumn{3}{c|}{Modality} & \multicolumn{1}{c|}{5-fold Acc} & \multicolumn{8}{c}{Category} \\
    \cmidrule(r){1-4}
    \cmidrule(r){5-13}
    & Visual & Audio & Object & $Acc. \pm std$ (\%) & C. Trait & R. Inter & Scene Id. & Story. U & Situ. U & Sent. & Audio & Mani.\\
    \midrule
    \multirow{6}{*}{Baseline} & CNN & - & - & $9.24 \pm 0.88$ & 13.97 & 5.73 & 9.63 & 6.01 & 8.45 & 7.30 & 17.34 & 2.76 \\
     & S3D & - & - & $11.60 \pm 0.76$ & 13.70 & 8.85 & 11.85 & \textbf{10.94} & 9.73 & 13.87 & 18.60 & 12.71 \\
     & - & Sound & - & $4.13 \pm 0.89$ & 7.95 & 3.12 & 1.48 & 3.54 & 2.57 & 6.93 & 10.57 & 4.97 \\
     & - & ASR & - & $5.44 \pm 0.79$ & 9.86 & 5.73 & 6.30 & 3.24 & 3.96 & 8.76 & 8.67 & 2.21 \\
     & S3D & Sound  & - & $10.48 \pm 1.17$ & 15.07 & 9.90 & 12.96 & 7.70 & 8.66 & 10.22 & 18.39 & 10.50 \\
     & S3D & ASR  & - & $12.01 \pm 0.90$ & 13.97 & 10.42 & 14.44 & 8.78 & 11.12 & 14.23 & 20.93 & 11.60 \\
    \midrule
    L-GCN \cite{L-GCN2020AAAI} & - & - & frcnn & $8.16 \pm 1.80$ & 9.04 & 6.77 & 11.48 & 5.24 & 8.34 & 6.93 & 10.15 & 8.84 \\
    \midrule
    \multicolumn{4}{l|}{XDC \cite{xdc}} & $7.35 \pm 0.52$ & 10.96 & 3.12 & 6.67 & 5.86 & 5.88 & 7.30 & 15.22 & 6.63 \\
    \midrule[1pt]
    \multirow{2}{*}{\ourmodelabbr{}} & S3D & ASR & - & $12.88 \pm 0.72$ & 17.26 & \textbf{19.79} & 9.63 & 10.63 & 10.91 & 17.52 & \textbf{26.00} & 7.18\\
     & S3D & ASR & frcnn & \textbf{13.94} $\pm 1.78$ & \textbf{19.73} & 16.67 & \textbf{15.93} & 10.32 & \textbf{13.69} & \textbf{18.98} & 20.72 & \textbf{13.26} \\
    \midrule[1pt]
    \multicolumn{4}{l|}{Oracle (w/ written description)} & $27.84 \pm 2.14$ & 40.55 & 31.77 & 31.48 & 17.41 & 26.42 & 30.29 & 32.98 & 21.55\\
    \midrule
    \bottomrule
  \end{tabular}
  \caption{The experimental result with 5-fold cross validation. First block: The baseline model with different modalities. The model using S3D and ASR reaches the highest score of 12.01\% accuracy. Second and third blocks are two state-of-the-art methods in Video QA and action recognition, they can achieve only 7$\sim$8\% accuracy. Fourth block: Our \ourmodelfull{} model: \ourmodelabbr{}, using raw signals (visual, audio, and objects) for conceptual storyteller and trope understanding reaches the best score at 13.94\%. Last block: Oracle case accessing human-written descriptions could achieve 27.84\% accuracy, which is remarkably better than all compared methods. (See Section \ref{section:experiment:results})
  }
  \label{tab:exp}
\end{table*}


%% file: Figures/ablation.tex
\begin{table}[]
    \centering
    \begin{tabular}{c|ccc}
    \toprule
         \multirow{2}{*}{Modality} & \multicolumn{3}{c}{Model}\\
          & Baseline & \ourmodelabbr{} (w/o ST.Emb) & \ourmodelabbr{} \\
         \midrule
         V+A & $12.01 \pm 0.90$ & $11.98 \pm 1.33$ & $12.88 \pm 0.72$ \\
         \midrule
         V+A+O & $12.50 \pm 0.52$ & $13.50 \pm 1.29$ & $13.94 \pm 1.78$ \\
    \bottomrule
    \end{tabular}
    \caption{
    Ablation study on \ourmodelabbr{}. The modality of V indicates the video features, A is ASR features, and O stands for object features. Three different types of models are baseline model, \ourmodelabbr{} without feeding story embedding back to trope understanding module (multitask structure only), and our full \ourmodelabbr{} model. The results show that multitasking alone can guide the video encoder to represent the video embedding better with adequate input signals (V+A+O). The full \ourmodelabbr{} model further improves about 0.9$\sim$1.5\% of accuracy to both of the variants.
    }
    \label{tab:ablation}
\end{table}

%% file: Figures/Cases.tex
\begin{figure*} 
\includegraphics[width=\textwidth]{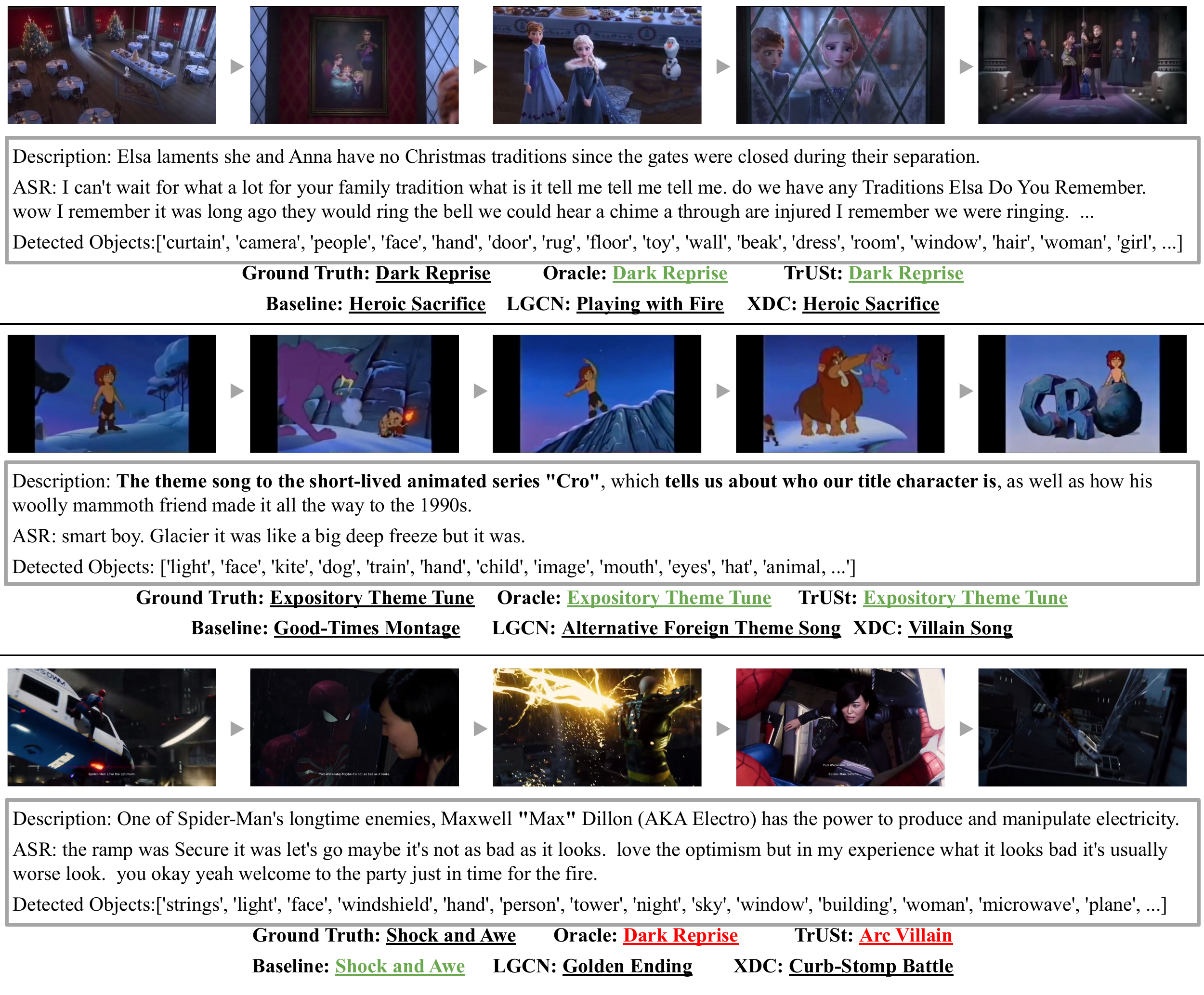}
\caption{
{\bfseries Qualitative results. The first two cases show the effectiveness of \ourmodelabbr{}. Our \ourmodelabbr{} benefits from the story embedding and understands the video implications without requiring human-written descriptions during inference. At the same time, other models can barely realize the shallow meaning of these videos and lead to the wrong answers. The third one is a failure case of the oracle model and \ourmodelabbr{}. The trope shows the weapon's special effects provided by the raw signals but not the descriptions. Despite generally performing better, our model failed in this case. (See Section \ref{para:analysis} for details.)
}
}
\label{fig:cases}
\end{figure*}